\documentclass[manuscript]{acmart}
\usepackage{booktabs}
\usepackage{multirow}
\AtBeginDocument{%
  \providecommand\BibTeX{{%
    \normalfont B\kern-0.5em{\scshape i\kern-0.25em b}\kern-0.8em\TeX}}}

\begin{document}

% \title[Mapping form Monomodal to Multimodal Behavioural Indicators]{Mapping Monomodal to Multimodal Behavioural Indicators in Collaborative Learning Analytics Using Latent Class Analysis}

\title[Integrating Latent Class Analysis to Uncover Multimodal Learning Patterns]{From Complexity to Parsimony: Integrating Latent Class Analysis to Uncover Multimodal Learning Patterns in Collaborative Learning}

\author{Lixiang Yan}
\affiliation{%
  \institution{Monash University}
  \country{Australia}
}

\author{Dragan Gašević}
\affiliation{%
  \institution{Monash University}
  \country{Australia}
}

\author{Linxuan Zhao}
\affiliation{%
  \institution{Monash University}
  \country{Australia}
}

\author{Vanessa Echeverria}
\affiliation{%
  \institution{Monash University}
  \country{Australia}
}

\author{Yueqiao Jin}
\affiliation{%
  \institution{Monash University}
  \country{Australia}
}

\author{Roberto Martinez-Maldonado}
\affiliation{%
  \institution{Monash University}
  \country{Australia}
}

\renewcommand{\shortauthors}{Yan et al.}
\begin{abstract}

Multimodal Learning Analytics (MMLA) leverages advanced sensing technologies and artificial intelligence to capture complex learning processes, but integrating diverse data sources into cohesive insights remains challenging. This study introduces a novel methodology for integrating latent class analysis (LCA) within MMLA to map monomodal behavioural indicators into parsimonious multimodal ones. Using a high-fidelity healthcare simulation context, we collected positional, audio, and physiological data, deriving 17 monomodal indicators. LCA identified four distinct latent classes: Collaborative Communication, Embodied Collaboration, Distant Interaction, and Solitary Engagement, each capturing unique monomodal patterns. Epistemic network analysis compared these multimodal indicators with the original monomodal indicators and found that the multimodal approach was more parsimonious while offering higher explanatory power regarding students' task and collaboration performances. The findings highlight the potential of LCA in simplifying the analysis of complex multimodal data while capturing nuanced, cross-modality behaviours, offering actionable insights for educators and enhancing the design of collaborative learning interventions. This study proposes a pathway for advancing MMLA, making it more parsimonious and manageable, and aligning with the principles of learner-centred education.

\end{abstract}



\maketitle

\section{Introduction}

Collaborative learning is a cornerstone of 21st-century education, encouraging students to solve problems, understand concepts, complete tasks, and achieve goals collectively \cite{laal2012benefits}. Understanding collaborative learning is complex, drawing on various learning and cognitive theories \cite{wise2021collaborative}. Traditional research has focused on cognitive, social cognitive, and social constructivist theories, examining communication processes, interaction patterns, and discourse in collaborative environments \cite{ferreira2021if}. Recent studies, especially in co-located physical settings, highlight the role of bodily behaviours. Grounded in the theory of situated and embodied cognition \cite{brown1989situated, wilson2002six}, these studies value physical interaction and digital resource use \cite{ioannou2019understanding}. In such settings, students' bodily and physiological behaviours provide critical insights into their learning processes and experiences \cite{ferreira2021if, yan2022assessment}. Thus, the need to study collaborative learning through a multimodal lens is increasingly recognised \cite{cukurova2020promise, yan2024evidence, spikol2017using}.

The rapid advancements in sensing technologies and artificial intelligence have significantly expanded the availability of diverse data sources to capture and understand complex learning processes, including collaborative learning \cite{wise2021collaborative, sharma2020multimodal, giannakos2019multimodal, ochoa2016augmenting}. Various modalities, such as digital, audio, positional, and physiological traces, provide a comprehensive representation of collaborative learning experiences \cite{worsley2021sixteen, yan2022scalability, crescenzi2020multimodal, cukurova2020promise}. This data richness is crucial for examining intricate learning behaviours and interactions, particularly where cognitive, metacognitive, and emotional states interplay \cite{sharma2020multimodal, ochoa2022multimodal, cukurova2020promise}. Capturing these detailed traces allows researchers to overcome the limitations of conventional data collection techniques \cite{cukurova2020promise, shum2012social, alwahaby2021}. Consequently, multimodal learning analytics (MMLA) has garnered increasing interest in the learning analytics and educational technology research community \cite{blikstein2013multimodal, yan2022scalability, alwahaby2021, wang2024data}. By integrating and analysing data from multiple modalities \cite{blikstein2013multimodal}, MMLA leverages digital, physical, and physiological data to provide enriched insights into dynamic learning activities \cite{ochoa2022multimodal}, informing targeted interventions to enhance educational outcomes \cite{cukurova2020promise, sharma2020multimodal, alwahaby2021}. MMLA studies have yielded novel insights into learning performance \cite{pijeira2016investigating, yan2022assessment}, stress levels \cite{ronda2021towards, yan2023physiological}, metacognitive processes \cite{dindar2020does, malmberg2021revealing}, socially shared regulation of learning \cite{noroozi2019multimodal, sobocinski2021exploring}, and social interactions \cite{chng2020using, saquib2018sensei} in collaborative environments.

Despite MMLA's potential, integrating different modalities to provide cohesive insights remains a significant challenge. Most existing data fusion studies have prioritised enhancing predictive learning analytics over enriching diagnostic learning analytics \cite{wang2024data, chango2022review, mu2020multimodal, samuelsen2019integrating}. While improved predictivity aids in developing early warning systems and decision-making \cite{sghir2023recent}, understanding the learning process is crucial for identifying reasons behind student performance \cite{hantoobi2021review}. Recent MMLA studies have often relied on monomodal behavioural indicators, focusing on single modalities like verbal communication \cite{zhao2023mets, southwell2022challenges, praharaj2021towards} or physiological synchrony \cite{dindar2020does, haataja2018monitoring, yan2023physio}, rather than synthesising multimodal indicators \cite{chango2022review, yan2022scalability, alwahaby2021, wang2024data}. Monomodal indicators offer insights into specific modalities (e.g., combining audio and positional data for verbal indicators \cite{zhao2023mets}), whereas multimodal indicators provide a holistic perspective by integrating insights from multiple modalities (e.g., combining verbal and gesture indicators for mathematical thinking). As MMLA's capability to capture various modalities grows, the complexity of interpreting the data increases \cite{yan2022scalability, zhao2024epistemic, dominguez2021scaling, jarvela2022multimodal}. Only initial efforts have been made to map monomodal to multimodal indicators \cite{giannakos2019multimodal, cukurova2019learning}, particularly for understanding the learning process. This limitation can hinder MMLA's effectiveness in capturing nuanced insights, supporting theory development, and facilitating stakeholder interpretation \cite{chango2022review, wang2024data}.

% The lack of methodologies for mapping monomodal to multimodal indicators can limit MMLA's effectiveness in capturing nuanced insights and hinder stakeholder interpretation \cite{chango2022review, wang2024data}.

This study addresses the critical challenge of integrating diverse data modalities within MMLA by introducing a novel approach to mapping monomodal to multimodal behavioural indicators through Latent Class Analysis (LCA). LCA, a statistical technique that identifies latent subgroups sharing similar patterns \cite{hagenaars2002applied}, is leveraged to uncover underlying data patterns across various modalities. This integrative methodology aims to streamline the identification and interpretation of theoretically and empirically valid multimodal indicators, thereby potentially enhancing our understanding of collaborative learning processes. To demonstrate the efficacy of this approach, we conducted a case study comparing 17 monomodal indicators with four resultant multimodal indicators, capturing key insights into students' collaborative experiences. The findings underscore the importance of developing interpretable, cohesive multimodal indicators that can optimise educational practices and improve stakeholder comprehension in MMLA. This study proposes a significant advancement in the field by reducing the complexity of MMLA models, making complex multimodal data more manageable and actionable for educational applications.
% making the interpretation of complex multimodal data more manageable and actionable for educational applications.

\vspace{-10pt}
\section{Background}

\subsection{Complexity of Collaborative Learning}

Collaborative learning is underpinned by multifaceted theoretical perspectives, notably social constructivism, situated cognition, and embodied cognition \cite{laal2012benefits, wise2021collaborative}. Grounded in Vygotsky's work, social constructivism posits that learning is an active, interactive process where students engage in knowledge construction through verbal exchanges and cooperative efforts, underlining the importance of social interactions in enabling tasks beyond individual capabilities \cite{vygotsky1978mind,miyake2007computer,mercer2004sociocultural}. Situated cognition extends this view, emphasising the context of learning, arguing that real-world environments enhance understanding and performance, thus valuing social interaction and cultural tools as essential to cognitive development \cite{brown1989situated,lave1991situated,gresalfi2011learning,johri2011situated}. Embodied cognition further highlights the role of bodily behaviours in learning, asserting that gestures, expressions, and movements are integral to cognitive processes, facilitating idea articulation and knowledge co-construction in collaborative settings \cite{wilson2002six,barsalou2008grounded,goldin2005hearing,gregorcic2017doing}. Understanding collaborative learning, therefore, requires integrating these theories to consider both verbal and non-verbal interactions, navigational movements, and the use of physical and digital resources \cite{ferreira2021if,zhao2022modelling,yan2022mapping,ioannou2019understanding}. However, capturing these rich, complex processes remains challenging.

\subsection{Multimodal Learning Analytics and Data Fusion}

Given the complexity of collaborative learning, traditional data collection methods, such as surveys, interviews, and direct observations, often fall short in capturing the full spectrum of student interactions, particularly their bodily behaviours \cite{luciano2018fitting, sobocinski2022exploring}. This limitation has led researchers to explore more robust and comprehensive approaches like MMLA. MMLA bridges the gap between learning analytics and advanced sensing technologies to collect, analyse, and interpret data from multiple modalities, providing a richer, more nuanced understanding of students' learning behaviours, and cognitive and emotional states \cite{sharma2020multimodal, blikstein2013multimodal, ochoa2022multimodal}.

Unlike traditional methods, MMLA leverages advanced technologies such as eye-tracking, position tracking, physiological sensing, computer vision, and natural language processing to capture fine-grained data \cite{crescenzi2020multimodal, cukurova2020promise, yan2022scalability}. For instance, eye-tracking glasses and biometric sensors have been used to understand children's learning processes and engagement \cite{crescenzi2020multimodal}. Research cataloguing various modalities and learning outcomes explored in MMLA from 2010 to 2018 highlights the ecological validity improvements brought about by these technologies \cite{chua2019technologies}. These studies use both controlled lab settings and authentic classroom environments to gather comprehensive data, enabling researchers to observe learning activities more precisely. Furthermore, advancements in AI techniques, such as gesture detection and speech recognition, enrich MMLA by automating the recording and analysis of students' behaviours. This automation proves invaluable for evaluating team-based tasks, assisting classroom management, and providing real-time feedback, enhancing teaching and learning \cite{sharma2020multimodal, worsley2018multimodal, giannakos2019multimodal, yan2022scalability}. By capturing multiple data streams simultaneously, MMLA can offer a holistic view of the learning process. This approach integrates theories of social constructivism, situated cognition, and embodied cognition, potentially offering deeper insights into how students interact, learn, and collaborate in complex environments \cite{ioannou2019understanding}.

To realise its potential, MMLA research has utilised various data fusion techniques to integrate data from multiple sources \cite{wang2024data, chango2022review, mu2020multimodal, samuelsen2019integrating}. Yet, most prior fusion studies have focused on enhancing predictive learning analytics. \citet{mu2020multimodal} highlighted that integrating multimodal data via "many-to-one" or "many-to-many" methods is more effective than single-mode data for predicting learning outcomes. For example, \citet{giannakos2019multimodal} and \citet{cukurova2019learning} demonstrated superior skill acquisition and tutor performance predictions using multimodal data fusion (e.g., eye-tracking and facial video) compared to monomodal approaches. \citet{chango2022review} categorised fusion techniques into feature-level (early fusion), decision-level (late fusion), and hybrid fusion. In early fusion, \citet{yue2019recognizing} enhanced data integration by selecting the best features and reducing dimensionality using principal component analysis and a Kolmogorov-Smirnov test. For decision fusion, \citet{monkaresi2016automated} improved accuracy by using individual base classifiers for each data channel and making final classifications based on the highest decision probability. In hybrid fusion, \citet{chango2021improving} combined feature and decision fusion by first merging features from various sources and then using ensemble techniques to integrate classifier decisions, boosting predictive performance. While these methodologies have significantly enhanced predictive accuracy, they provide limited insights into the underlying learning processes.

Recent MMLA studies focusing on the learning process have often utilised monomodal rather than truly integrated multimodal behavioural indicators. For instance, \citet{xu2023multimodal} used audio, video, and digital trace data to derive separate indicators for verbal communication, facial expression, and operational behaviour in pair programming. Similarly, \citet{yan2024dissecting} transformed positional, audio, and heart rate data into distinct indicators of team prioritisation, communication, and engagement in healthcare simulations. While these studies provided valuable insights into collaborative patterns and identified behavioural distinctions between performance levels, each indicator remained modality-specific, with no data fusion to integrate these insights into cohesive multimodal indicators. This approach often leads to complex analytics due to the overwhelming number of monomodal indicators. Additionally, even though \citet{zhao2023mets} attempted early fusion of audio and positional data, the resulting indicators focused solely on verbal communication, lacking cross-modality insights. Consequently, there is a notable gap in methodologies for mapping monomodal indicators, which provide single-modality insights, to multimodal indicators that capture cross-modality interactions \cite{wang2024data, chango2022review}. Addressing this limitation is essential for MMLA to offer a holistic view of collaborative learning processes \cite{crescenzi2020multimodal, cukurova2020promise, yan2022scalability}.

\vspace{-10pt}
\subsection{Latent Class Analysis}

Latent Class Analysis (LCA) is a well-established method in learning analytics and educational data mining research, built on several key assumptions \cite{whitelock2021exploratory, viberg2022exploring, kang2020heterogeneity}. First, each person in a sample at a given time belongs to one and only one latent class, making these classes exhaustive and mutually exclusive. Second, individuals within a given class exhibit intragroup homogeneity, sharing similar behaviour patterns that distinguish them from other classes. Unlike traditional clustering methods that focus on grouping variables, LCA categorises individuals based on their behaviour patterns. The main objectives of LCA are to identify and enumerate the latent classes, characterise these classes based on their influence on observed behaviours, and classify individuals into these groups according to their behaviour patterns \cite{hagenaars2002applied}. Prior learning analytics studies have predominantly used this person-centred approach to understand individual differences in learners' experiences and perceptions by identifying latent classes based on survey responses \cite{whitelock2021exploratory, viberg2022exploring}. For example, one study used LCA to identify distinct subgroups of learners' behavioural patterns in a MOOC by transforming their digital traces into a sequence of categorical data, assigning values like 1 for completing a video or quiz and 0 if not, resulting in five latent classes with unique combinations of video viewing and quiz completion behaviours \cite{kang2020heterogeneity}. Compared to other data compression methods, such as principal component analysis, factor analysis, and singular value decomposition, LCA offers greater explainability. This is because LCA retains lower-level insights, with each latent class encompassing unique behavioural patterns \cite{hagenaars2002applied}. Despite its advanced statistical capabilities for uncovering behaviour patterns, LCA is rarely applied in MMLA research to identify latent classes of multimodal behavioural patterns from monomodal indicators.

\vspace{-10pt}
\subsection{Research Questions}

This study aims to address the limitation of monomodal indicator reliance by proposing a methodology that integrates LCA into MMLA. This methodology transforms monomodal indicators into multimodal indicators that encompass cross-modality insights. The objective is to uncover hidden patterns and relationships across different data modalities, leading to a more comprehensive understanding of student collaboration, communication, and engagement in collaborative learning environments. To demonstrate the efficacy of the proposed methodology, we utilised a robust dataset comprising various behavioural indicators collected from multiple modalities such as audio, positional, and physiological data. Specifically, the following research questions were investigated to illustrate the methodology and validate the effectiveness of multimodal indicators in capturing cross-modality insights: \textbf{RQ1}: How can LCA be utilised within MMLA to derive multimodal behavioural indicators from monomodal behavioural indicators? \textbf{RQ2}: What are the differences in behavioural patterns of students with varying levels of learning experience when analysed through multimodal behavioural indicators compared to monomodal behavioural indicators?

\section{Method}
\label{sec-method}

In this section, we first introduce the learning setting of our illustrative example, which is used to demonstrate and assess our proposed methodology. Next, we elaborate on the proposed methodology through the lens of contemporary MMLA models and data fusion methodologies, highlighting the added value of incorporating LCA in MMLA with our example. Finally, we outline the analysis conducted to investigate the two research questions, aiming to understand the efficacy of multimodal behavioural indicators in providing insights into collaborative learning processes.

\subsection{Learning Setting and Data}
\label{sec-method-context}

The learning setting present in this study is set in a high-fidelity healthcare simulation environment, a widely used instructional method in healthcare education that replicates real clinical situations with sophisticated patient manikins \cite{theodoulou2018simulation}. Aligning with constructivist learning theories, this technique engages students in active, reflective tasks within authentic contexts \cite{lateef2010simulation}. The simulation classroom featured four patient beds, medical equipment, and advanced patient manikins controlled by teaching staff. Overseen by experienced educators, the main aim was to provide students with opportunities to practice teamwork, communication, and prioritisation skills in a realistic setting. In each session, four students managed the beds, focusing primarily on a deteriorating patient (Bed 4) while handling secondary tasks and a distracting teaching staff member acting as a relative (Bed 3). The activity had four stages, with students initially briefed by staff and then free to strategise as the patient in Bed 4 showed signs of deterioration. The third and fourth stages involved all four students working together, making it the focal point for analysing dynamic, collaborative behaviours and task performance variations.

Three types of behavioural modalities were collected using wearable sensing technologies, along with students' self-reported experiences. Positioning data were recorded with the Pozyx creator toolkit, which employed Ultra-Wideband (UWB) tags and anchors to determine real-time x-y coordinates. Audio data were captured using compact wireless headset microphones (Xiaokoa) and synchronised through a multi-channel audio interface (TASCAM US-16×08). Physiological data, including heart rate, electrodermal activity (EDA), and blood volume pulse (BVP), were collected using Empatica E4 wristbands. Due to the low quality of EDA data, likely caused by electrode-skin disconnections during physical activity, our analysis focused on heart rate. The heart rate, calculated automatically from BVP data at 1 Hz by the Empatica device, demonstrated high reliability with a median inter-device correlation of 0.94 and a median intraclass correlation coefficient of 0.94 \cite{van2022ambulatory}. We successfully captured multimodal data from 56 students across 14 simulation sessions. Additionally, students' collaborative learning experiences were measured using two self-reported surveys rated on a seven-point Likert scale, ranging from 1 (\textit{strongly disagree}) to 7 (\textit{strongly agree}). These surveys assessed satisfaction with task performance (\textit{"I am satisfied with my task performance during the simulation"}; M = 4.53, SD = 1.25) and satisfaction with team collaboration (\textit{"I am satisfied with the performance of my group in terms of collaboration"}; M = 5.61, SD = 1.07). These single-item measures of satisfaction have demonstrated high reliability \cite{wanous1997overall}.

\subsection{Proposed Methodology: Mapping Multimodal Indicators}

\subsubsection{From theory to monomodal indicators}
\begin{figure}[htbp]
    \centering
    \includegraphics[width=.9\textwidth]{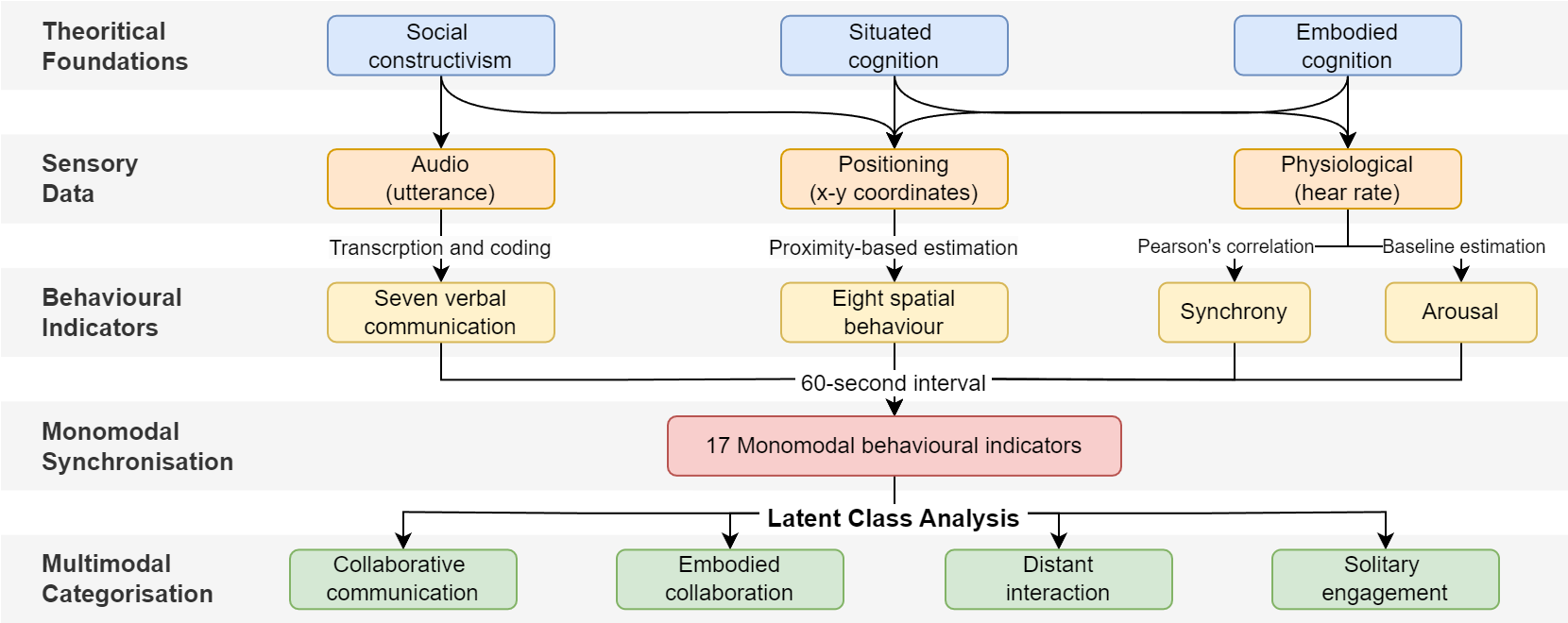}
    \caption{Mapping from sensory data to monomodal behavioural codes and then to multimodal behavioural codes.}
    \label{fig-procedure}
\end{figure}

The proposed methodology builds on contemporary MMLA models that map sensory data to monomodal behavioural indicators. Worsley et al. \cite{worsley2016situating} introduced a model illustrating the relationships between learning constructs, behavioural indicators, analytic techniques, tools, data, and sensors. This backward mapping model visually represents the decision-making process for selecting relevant sensors to capture specific learning constructs. Echeverria \cite{echeverria2020designing} later refined this model by incorporating human-centred design principles, an inductive backward mapping process, and a theory-driven approach, thereby creating a systematic structure to contextualise fine-grained sensor data in nursing simulations. More recently, Ochoa \cite{ochoa2022multimodal} enhanced the model by adding an execution layer that outlines the methodological procedures for developing MMLA systems, including data collection, feature extraction, learning analytics, and educational insights. Consequently, the mapping from sensory data to monomodal behavioural indicators follows the steps outlined in these prior MMLA models, where learning and cognitive theory motivate the utilisation of different sensory data, which can be transformed into monomodal indicators that offer insights into specific behavioural modalities (Figure \ref{fig-procedure}).

As shown in Figure \ref{fig-procedure}, for the current example, the sensor data collected were transformed into three types of monomodal behavioural indicators. Specifically, positioning data were used to identify task prioritisation behaviours, audio data were transcribed to capture team communication behaviours, and heart rate data were analysed to assess physiological arousal and synchrony.

\paragraph{Eight task prioritisation indicators.} Developing task prioritisation skills is crucial for healthcare students \cite{tiwari2005student}. The simulation room was divided into distinct zones for specific tasks, based on learning design and expert advice. Each bed had a designated centroid with a 1.5-metre radius \cite{hall1971proxemics}, marking primary and secondary task areas. These zones were mutually exclusive, and collaboration was identified when two students were within one metre of each other for more than 10 seconds. We defined four prioritisation categories based on students' spatial behaviours (SP): \textit{SP.collaborate.primary} and \textit{SP.individual.primary} for primary task spaces, \textit{SP.collaborate.secondary} and \textit{SP.individual.secondary} for secondary task spaces, and \textit{SP.collaborate.distraction} and \textit{SP.individual.distraction} for the distraction task space. Additionally, behaviours outside the predefined task zones were classified as \textit{SP.task.distribution} when two or more students were co-located outside task zones for a sustained period, suggesting discussions about task responsibilities, and \textit{SP.task.transition} when a student was outside task zones without being near other students. This approach allowed us to capture nuanced task prioritisation and collaboration dynamics from positioning data \cite{yan2022assessment}.

\paragraph{Seven team communication indicators.} The audio data were manually transcribed and coded based on a scheme adapted from previous research on healthcare teamwork and collaborative learning \cite{miller2009identifying, barton2018teaching, bigfive}. This coding scheme included four main constructs and seven specific verbal communication behaviours (VB), with inter-rater reliability assessed using Cohen's kappa, achieving satisfactory agreement for each code \cite{mchugh2012interrater}. The construct of \textbf{Shared Leadership} covers \textit{VB.task.allocation}, involving the explicit assignment of tasks or polite inquiries (Cohen's kappa = 0.80), and \textit{VB.handover.provision}, which follows specific handover communication protocols (Cohen's kappa = 0.78). \textbf{Situation Awareness} includes \textit{VB.escalation}, occurring when a student identifies the need for additional assistance, such as calling for a doctor (Cohen's kappa = 0.85). \textbf{Shared Mental Model} encompasses \textit{VB.information.sharing}, where relevant information is proactively shared (Cohen's kappa = 0.74); \textit{VB.information.requesting}, involving the seeking of information from others (Cohen's kappa = 0.86); and \textit{VB.responding.to.requests}, which pertains to providing information when asked (Cohen's kappa = 0.80). Lastly, \textbf{Closed-loop Communication} includes \textit{VB.agreement}, where communication is finalised through explicit verbal affirmations (Cohen's kappa = 0.86). This thorough coding process enabled us to derive detailed communication features from the audio data, shedding light on the dynamics of collaborative learning.

\paragraph{Two physiological indicators.} Heart rate data were utilised to model two physiological indicators (PY). \textit{PY.synchrony} was calculated by computing Pearson's correlation among all student pairs in a simulation, which was then averaged for each student \cite{gordon2021group}. \textit{PY.arousal} was determined when a student's heart rate exceeded their baseline, which was defined as the average heart rate during the first phase (limited physical movement and verbal acitivity).

\subsubsection{Synchronising Monomodal Behavioural Indicators}

To derive multimodal behavioural indicators that capture cross-modality insights from various monomodal indicators, it is crucial to first synchronise these indicators into a uniform unit of analysis. Given the sequential and temporal nature of behaviours in a collaborative learning setting, we used an established interleaving approach, which examines the temporal interplay between events recorded across different modalities \cite{sung2022methods}. This method interleaves different monomodal indicators within a single dataset, allowing us to model moment-by-moment patterns across multiple modalities and track the temporal evolution of these patterns throughout the collaborative learning process.

\begin{figure}[htbp]
    \centering
    \includegraphics[width=.99\textwidth]{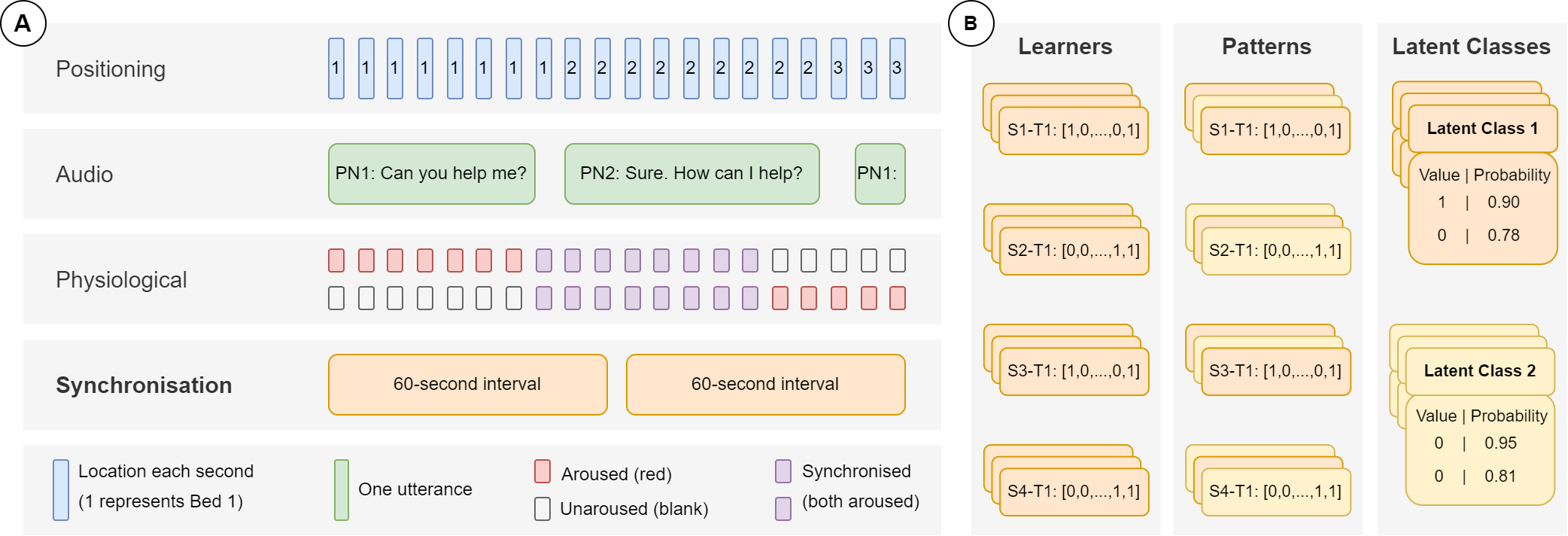}
    \caption{Illustration of A) synchronising positioning, audio, and physiological data into 60-second intervals and B) categorising each learner (e.g., S1) at a given 60-second interval (e.g., T1) into a distinct latent class based on their behaviour patterns (e.g., [1,0,...,0,1]).}
    \label{fig-synchronisation}
\end{figure}

As depicted in Figure \ref{fig-synchronisation} (A), each monomodal indicator is recorded at different frequencies. Task prioritisation (positioning) and physiological indicators were captured consistently at one-second intervals, whereas team communication indicators, derived from audio data, varied with utterance length. To establish a uniform unit of analysis, we aggregated the data to the largest granularity present—team communication indicators \cite{knight2017time}. It is crucial to select a unit of analysis that balances the weight of each monomodal indicator \cite{sung2022methods}. For example, a shorter interval (e.g., 10 seconds) could undervalue team communication indicators due to longer gaps between utterances, thus giving disproportionate weight to task prioritisation and physiological indicators captured every second. Consequently, we chose a 60-second interval for this learning context to effectively capture both verbal and nonverbal activities \cite{chejara2023impact}, such as a student requesting help with an oxygen mask and subsequently sharing the patient's blood oxygen levels. 

For a given 60-second interval, task prioritisation indicators were considered present (1) if the student engaged in the behaviour (e.g., \textit{SP.collaborate.primary}) for at least ten consecutive seconds, minimising misidentification from merely walking past a task space \cite{greenberg2014dark}. A team communication indicator (e.g., \textit{VB.information.sharing}) was considered present if the behaviour occurred at least once within the interval. Lastly, a physiological indicator was considered present if it was demonstrated for more than half of the interval (e.g., over 30 seconds) \cite{dindar2020does,malmberg2021revealing}. The resulting synchronised dataset for each 60-second interval is a sequence of 1s and 0s, where each 1 or 0 represents the presence or absence of a monomodal behavioural indicator (e.g., [1, 0,..., 1, 0]).

\subsubsection{Categorising Multimodal Behavioural Indicators}

Once the monomodal indicators are synchronised into consistent time intervals for each individual, LCA can be applied to identify distinct behavioural patterns across all modalities, provided all the necessary assumptions are met. Specifically, at any given time interval, each individual demonstrates a specific pattern of behaviours, thereby satisfying LCA's first assumption that each person belongs to one and only one latent class at a time. Although individuals' behaviours can transition from one latent class to another, such changes occur at different time points. This ensures the intragroup homogeneity for each latent class during any specific interval, thus fulfilling LCA's second assumption. Furthermore, by grouping monomodal indicators into sequences for each interval, we categorise the behaviour of individuals at each time point. This approach aligns with LCA's objective of categorising individuals based on observed behaviour patterns rather than the behaviour indicators themselves \cite{hagenaars2002applied}. Applying LCA in this manner allows us to uncover nuanced behavioural patterns in the collaborative learning process, providing a comprehensive understanding of how different modalities interact and evolve over time.

As shown in Figure \ref{fig-synchronisation} (B), for the current learning context, each learner (e.g., S1, S2, and S3) will have multiple segments of monomodal indicator sequences, each representing a specific 60-second interval in temporal and sequential order (e.g., T1, T2, and T3). This synchronised dataset can then be fed into the LCA model, which assumes the observed data can be explained by a finite number of latent classes, each representing a distinct behavioural pattern. To determine the optimal number of latent classes, we selected the LCA model with the lowest Bayesian Information Criterion (BIC) and the highest log-likelihood value \cite{hagenaars2002applied}. The LCA model then estimates the probability of each individual belonging to each latent class at different time intervals based on their behavioural sequences. Once the model parameters are estimated, we can interpret the latent classes by examining the probability profiles for each class, thereby identifying the distinct multimodal behavioural patterns that exist across the dataset. Individuals are then categorised into these latent classes, allowing us to track how their multimodal behaviours change over time within the collaborative learning setting. This categorisation enables the analysis of how different monomodal behavioural indicators co-occur and evolve over time. By understanding these patterns, cross-modality insights can be obtained, providing a comprehensive view of the interactions and behaviours in collaborative learning.

\subsection{Data Analysis}

To address RQ1, we applied the aforementioned methodology to identify distinct latent classes of multimodal behavioural indicators from the 17 monomodal indicators, including eight related to task prioritisation, seven to team communication, and two to physiology. We first performed a correlation analysis on all monomodal indicators to identify and eliminate highly correlated pairs (e.g., Spearman's \(\rho\) > 0.8), which can complicate model estimation and potentially hinder convergence \cite{lanza2013latent}. However, none of the pairs were highly correlated and thus, all the 17 monomodal indicators were used for LCA. We constructed LCA models ranging from one to ten classes and calculated the Bayesian Information Criterion (BIC) and log-likelihood to determine the optimal number of latent classes \cite{hagenaars2002applied}. For each latent class, we estimated the probability of each monomodal indicator for each student and assigned a value of 1 for the presence and 0 for the absence of these indicators. Line graphs were then created for each latent class, with the 17 monomodal indicators on the x-axis and presence (1) or absence (0) on the y-axis, to visualise the patterns of monomodal indicators. These distinct latent classes were then labelled based on the learning context to describe the multimodal indicators.

For RQ2, we used epistemic network analysis (ENA) to compare the co-occurrence of behaviour patterns captured by the 17 monomodal indicators with the identified multimodal indicators. ENA is a well-established method in learning analytics research for investigating the co-occurrence of behavioural, cognitive, and metacognitive patterns \cite{shaffer2016tutorial, csanadi2018coding, yan2023sena}. Specifically, we used the ENA Web Tool \cite{ENAwebtool} to construct epistemic networks for students satisfied (Likert scale: four or above) and dissatisfied (below four) with their self-reported task performance and collaboration. These dichotomous groups are commonly used in prior ENA studies to model differences in learning behaviours \cite{fan2022dissecting, yan2022assessment, zhao2023mets}. We applied a nested structure to account for group variance, with each student as the unit of analysis, each 60-second interval as the line level, and the entire learning activity (when all four students were present) as the stanza size. This approach is justified as students' behaviours throughout the learning activity are interrelated; for example, task completion at the start precludes performing the same task later. We used Means Rotation (MR) for dimensional reduction to optimise the differences between the mean of the two groups along the X-axis \cite{bowman2021mathematical}. This approach simplifies the interpretation of resulting networks, particularly for distinguishing group differences. Mann-Whitney \textit{U} tests were performed to examine statistical differences in task and collaboration satisfaction along the X and Y axes of the comparison plots. Bonferroni correction was applied for multiple comparisons, with an initial alpha value of 0.05.
\vspace{-10pt}
\section{Results}

\subsection{RQ1 -- Latent Classes of Multimodal Behaviours}

\begin{figure}
    \centering
    \includegraphics[width=1\linewidth]{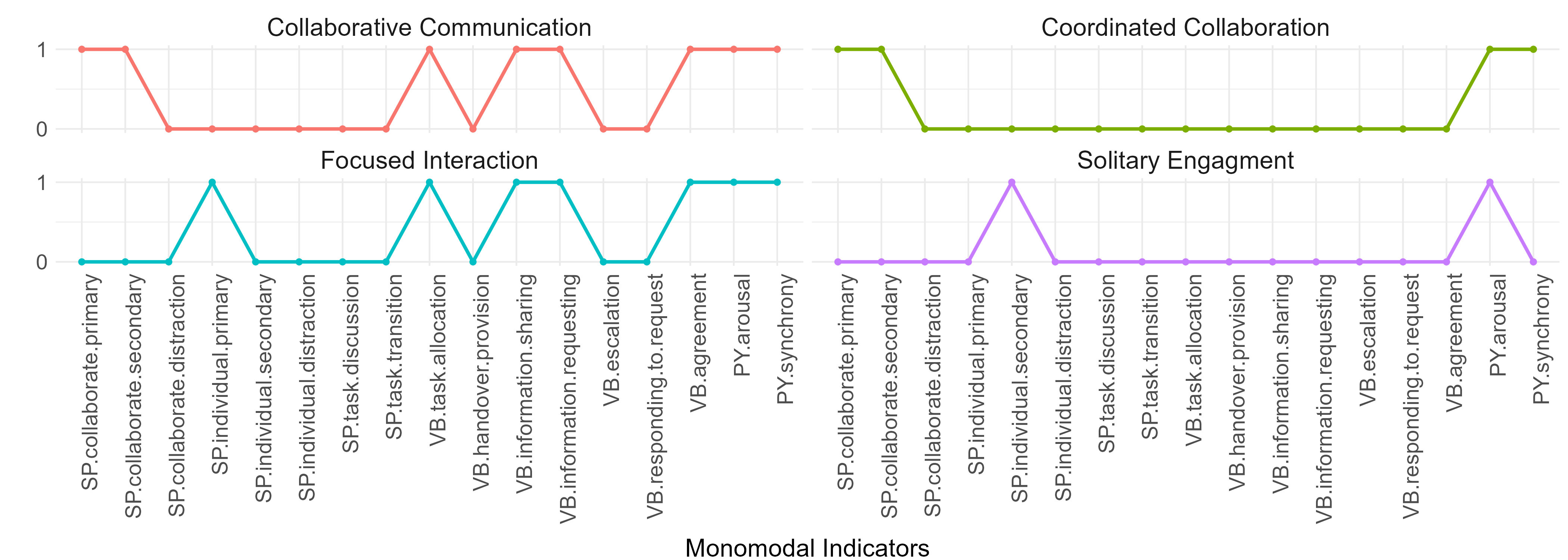}
    \caption{Four latent classes of multimodal behaviours (1 - present; 0 - absent), including Collaborative Communication (top-left), Embodied Collaboration (top-right), Distant Interaction (bottom-left), and Solitary Engagement (bottom-right).}
    \label{fig-rq1}
\end{figure}

Figure \ref{fig-rq1} depicts four distinctive latent classes of multimodal behavioural indicators identified through LCA, which provide cross-modality insights. The first latent class (Figure \ref{fig-rq1}; top-left) involves the presence of all three behavioural categories: task prioritisation, team communication, and physiological behaviours. Students demonstrating this behaviour in a given 60-second interval were working collaboratively on both the primary and secondary tasks (\textit{SP.collaborate.primary} and \textit{SP.collaborate.secondary}). They engaged in team communication by distributing task responsibilities (\textit{VB.task.allocation}), exchanging information (\textit{VB.information.sharing} and \textit{VB.information.requesting}), and participating in closed-loop communication (\textit{VB.agreement}). These students were both physiologically aroused (\textit{PY.arousal}) and synchronised (\textit{PY.synchrony}) with others during this interval. Consequently, we labelled this multimodal indicator as \textit{Collaborative Communication} based on the observed patterns.

The second latent class (Figure \ref{fig-rq1}; top-right) shares a similar pattern of task prioritisation and physiological behaviour with the first class. Students worked collaboratively on both the primary and secondary tasks (\textit{SP.collaborate.primary} and \textit{SP.collaborate.secondary}) and were both physiologically aroused (\textit{PY.arousal}) and synchronised (\textit{PY.synchrony}) with others. However, they did not engage in team communication behaviours, suggesting they might have been focused on procedural tasks or supporting others in specific activities (e.g., putting on an oxygen mask or preparing medications). We labelled this class as \textit{Embodied Collaboration}, highlighting moments when students demonstrated mostly embodied but not verbal behaviours, potentially playing a more supportive and task execution role.

The third latent class (Figure \ref{fig-rq1}; bottom-left) involves students focusing on working individually on the primary task (\textit{SP.individual.primary}). Despite this individual focus, they still engaged in substantial team communication behaviours such as distributing responsibilities (\textit{VB.task.allocation}), exchanging information (\textit{VB.information.sharing} and \textit{VB.information.requesting}), and participating in closed-loop communication (\textit{VB.agreement}). These students were also physiologically aroused and synchronised (\textit{PY.arousal} and \textit{PY.synchrony}) with the team, indicating their continued involvement in the team collaboration process but over a distant. We labelled this class as \textit{Distant Interaction}.

The final latent class (Figure \ref{fig-rq1}; bottom-right) involves students working individually on the secondary task (\textit{SP.individual.secondary}) while being physiologically aroused (\textit{PY.arousal}) but not synchronised. Unlike the other three classes, these students were neither collaborating nor communicating with others. Instead, they were engaged in solitary activities related to the secondary tasks. This multimodal indicator was labelled as \textit{Solitary Engagement}.

These findings indicate that LCA could effectively identify and categorise patterns of multimodal behaviours from individual monomodal indicators in MMLA. This categorisation resulted in four distinct latent classes of multimodal behavioural indicators: \textit{Collaborative Communication}, \textit{Embodied Collaboration}, \textit{Distant Interaction}, and \textit{Solitary Engagement}. These classes capture the different ways in which students interact and collaborate, providing valuable insights into team dynamics and individual contributions.

\subsection{RQ2 -- Difference in Post-task Measures}

\subsubsection{Task Performance}

\begin{figure}
    \centering
    \includegraphics[width=1\linewidth]{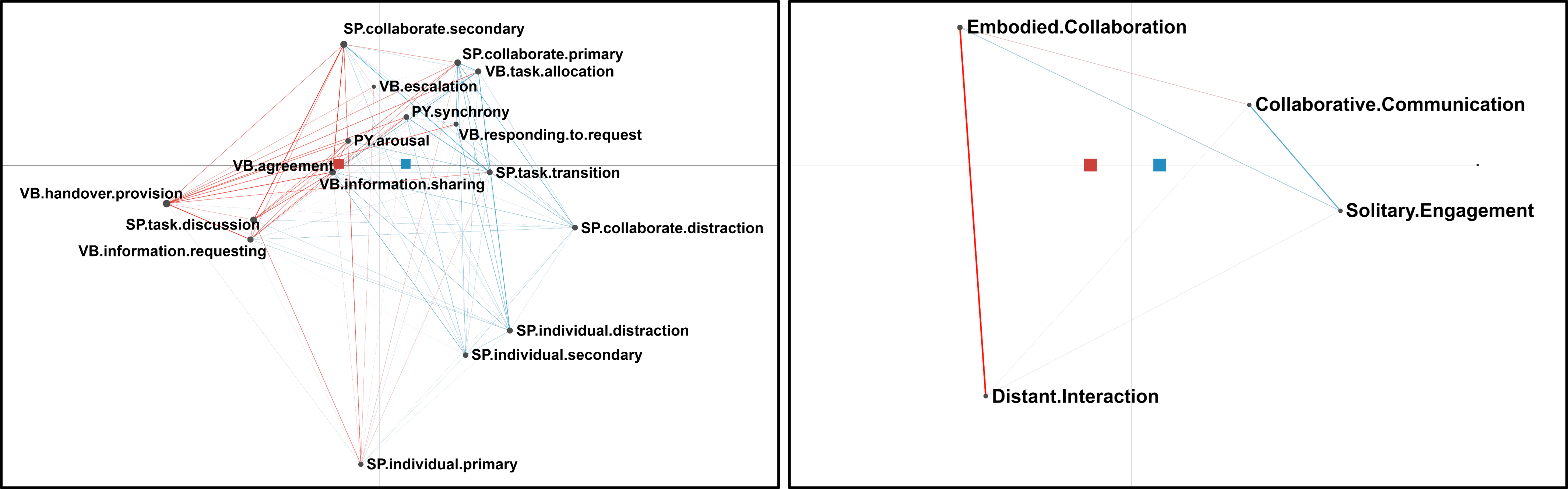}
    \caption{Comparison of epistemic networks generated from 17 monomodal (left) and four multimodal behavioural codes (right) for students with low (red) and high (blue) task performance satisfaction.}
    \label{fig-rq2-task}
\end{figure}

Regarding students' satisfaction with their task performance, the 17 monomodal indicators network (Figure \ref{fig-rq2-task}; left) revealed that unsatisfied students (Mdn=0.09, N=23) were statistically significantly different from satisfied students (Mdn=-0.09, N=33 U=231, p=0.014, r=0.39) along the x-axis (MR1; 9.3\% variance explained) but not along the y-axis (SVD2; 19.6\% variance explained; unsatisfied: Mdn=0.04, N=23; satisfied: Mdn=0.05, N=33 U=385, p=0.93, r=-0.01). Likewise, the epistemic network computed from the four multimodal indicators (Figure \ref{fig-rq2-task}; right) also revealed that unsatisfied students (Mdn=0.01, N=23) were statistically significantly different from satisfied students (Mdn=0.01, N=33 U=260, p=0.045, r=0.31) along the x-axis (MR1; 17.5\% variance explained) but not along the y-axis (SVD2; 44.5\% variance explained; unsatisfied: Mdn=-0.24, N=23; satisfied: Mdn=-0.24, N=33 U=382, p=0.97, r=-0.01). While both models identified significant differences along the x-axis, the model using four multimodal indicators (17.5\%) explained more variance along this axis than the model using 17 monomodal indicators (9.3\%).

\subsubsection{Collaboration Performance}

\begin{figure}
    \centering
    \includegraphics[width=1\linewidth]{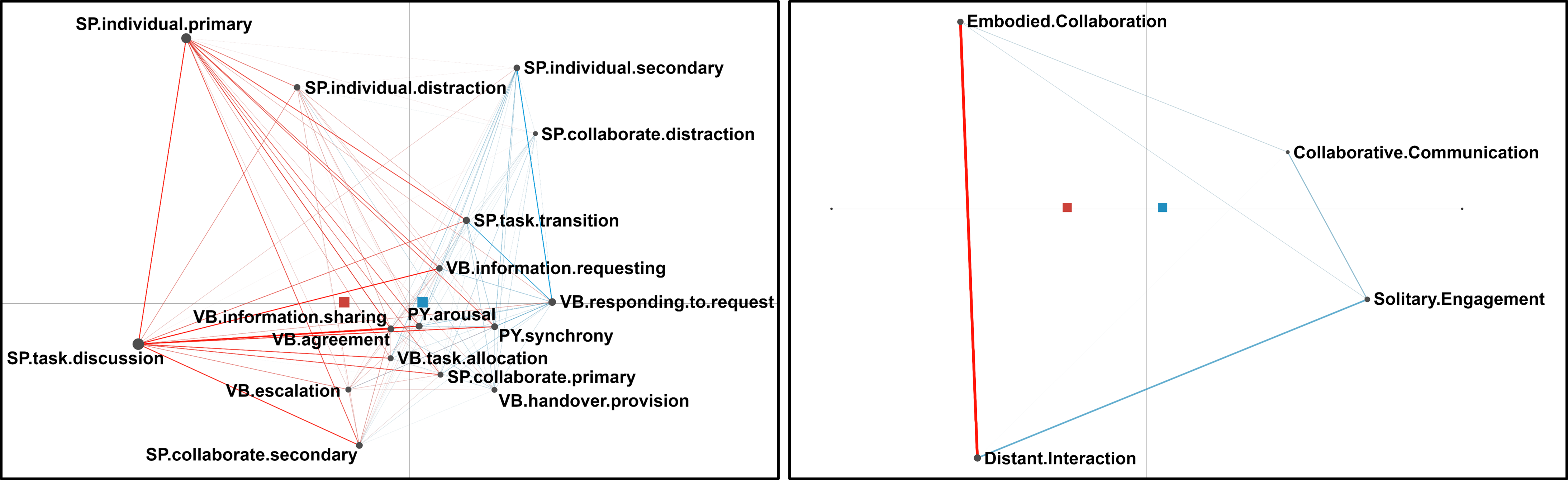}
    \caption{Comparison of epistemic networks generated from 17 monomodal (left) and four multimodal behavioural codes (right) for students with low (red) and high (blue) collaboration performance satisfaction.}
    \label{fig-rq2-collaboration}
\end{figure}

For students' satisfaction with their collaboration performance, the epistemic network constructed using the 17 monomodal indicators (Figure \ref{fig-rq2-collaboration}; left) revealed that unsatisfied students (Mdn=-0.10, N=10) were statistically significantly different from satisfied students (Mdn=0.00, N=46, U=94, p=0.004, r=0.59) along the x-axis (MR1; 8.5\% variance explained) but not along the y-axis (SVD2; 23.5\% variance explained; unsatisfied: Mdn=0.00, N=10; satisfied: Mdn=0.01, N=46, U=225, p=0.92, r=0.02). Similarly, the epistemic network computed from the four multimodal indicators (Figure \ref{fig-rq2-collaboration}; right) demonstrated that unsatisfied students (Mdn=-0.28, N=10) were statistically significantly different from satisfied students (Mdn=0.10, N=46, U=347, p=0.011, r=-0.51) along the x-axis (MR1; 15.6\% variance explained) but not along the y-axis (SVD2; 45.9\% variance explained; unsatisfied: Mdn=0.030, N=10; satisfied: Mdn=0.03, N=46, U=223, p=0.90, r=0.03). Once again, both models captured statistical differences along the x-axis, but the four multimodal indicators model (15.6\%) explained more variance on this x-axis than the 17 monomodal indicators model (8.5\%). Therefore, the four multimodal indicators model is not only more parsimonious but also has more explanatory power in accounting for differences between students' satisfaction with their tasks and collaboration performance.

\subsubsection{Co-occurrence Interpretation}

The comparison plots computed from the 17 monomodal indicators for both task satisfaction (Figure \ref{fig-rq2-task}; left) and collaboration satisfaction (Figure \ref{fig-rq2-collaboration}; left) are complex and difficult to interpret due to the intricate patterns of co-occurrence among different monomodal indicators. Specifically, although we can identify several iconic monomodal indicators for unsatisfied (e.g., \textit{VB.handover.provision} and \textit{SP.task.discussion} for task satisfaction; \textit{SP.task.discussion} and \textit{SP.individual.primary} for collaboration satisfaction) and satisfied students (e.g., \textit{SP.collaborate.distraction} and \textit{SP.individual.distraction} for task satisfaction; \textit{VB.responding.to.request} and \textit{SP.individual.secondary} for collaboration satisfaction), it is challenging to understanding the occurrence of these monomodal indicators and relate and rationalise them through the lens of the learning context. Additionally, it is also difficult to remove any monomodal indicators as none of the pairs were highly correlated (e.g., Spearman's \(\rho\) > 0.8), suggesting that they each provide unique information for understanding the collaboration process. 

On the other hand, the four multimodal indicator models were simpler and easier to interpret. Specifically, students who were unsatisfied with both their task and collaboration performances were characterised by the strong co-occurrence of \textit{Embodied Collaboration} and \textit{Distant Interaction}. Whereas students satisfied with their task performance were characterised by the co-occurrence of \textit{Collaborative Communication} and \textit{Solitary Engagement}. Likewise, students satisfied with their collaboration performance were characterised by the co-occurrences that \textit{Solitary Engagement} had with \textit{Distant Interaction} and \textit{Collaborative Communication}. These findings indicate that students' dissatisfaction with their task and collaboration performances seems to have been related to the transition between working individually on the primary task while engaging in distant verbal communication with others (\textit{Distant Interaction}) and working collaboratively on the primary and secondary tasks with little verbal communication (\textit{Embodied Collaboration}). A potential explanation is that these frequent transitions may create disruption and inefficiencies in the workflow, leading to decreased satisfaction. However, the direction of such transitions remains unknown from ENA. 
\section{Discussion and Conclusion}

This study introduced and demonstrated an innovative approach integrating LCA within MMLA, effectively bridging monomodal and multimodal indicators for richer insights. In RQ1, we showcased LCA's efficacy in identifying four distinct multimodal indicators from 17 monomodal indicators that encompass positional, verbal communication, and physiological data. Rather than separately analysing the relationships between monomodal indicators, each resultant multimodal indicator from the LCA highlighted unique patterns among the 17 monomodal indicators. This cross-modality perspective provided a comprehensive view of behavioural co-occurrences within a given 60-second interval of the collaborative learning activity. The findings align with LCA's established utility in prior studies, which have explored students' experiences and behaviours through survey instruments and digital traces \cite{whitelock2021exploratory, viberg2022exploring, hagenaars2002applied}. Moreover, our study extends the applicability of LCA within learning analytics to multimodal data, refining monomodal indicators into a more manageable set of informative multimodal indicators. By identifying latent classes, we are also effectively recognising underlying constructs, similar to approaches used in statistical models with surveys and factor analysis, facilitating the transition from data to theory building and validation and addressing a common challenge in learning analytics \cite{knight2017theory}. This approach could take MMLA closer to its goal of delivering evidence-based theories and enriched insights into dynamic learning activities by integrating learners' digital, physical, and physiological data \cite{ochoa2022multimodal, cukurova2020promise, blikstein2013multimodal}.

Regarding our second research question, while the 17 monomodal indicators provided granular insights into students' behavioural patterns based on their learning experience, the complexity of analysing and interpreting these relationships increased with the number of indicators. This echoes previous concerns that as MMLA’s capability to capture different modalities grows, so too does the complexity of interpreting the resulting data \cite{yan2022scalability, zhao2024epistemic, dominguez2021scaling, jarvela2022multimodal}. Contrarily, utilising LCA to categorise monomodal indicators into multimodal ones offers a potential solution to manage this complexity without significantly losing information. Our findings demonstrated that the four multimodal indicators had higher variance explained compared to the 17 monomodal indicators when conducting ENA. This is consistent with LCA being a person-centred rather than a variable-centred approach \cite{hagenaars2002applied}, wherein each learner can be classified by a specific multimodal indicator at a given time. These indicators still retain the insights from the original monomodal data, displaying distinct monomodal patterns. Consequently, this approach provides dual levels of insights: higher-level cross-modality behaviours contextualised within the learning environment and more detailed, fine-grained insights into each specific modality.

The current findings have several implications for future MMLA research and practice. Firstly, incorporating LCA within MMLA can streamline data analysis and interpretation, making it more manageable while preserving data granularity. Researchers and practitioners can gain a holistic perspective on collaborative learning behaviours, useful for designing and implementing interventions \cite{ferreira2021if}. Educators can tailor instructional strategies based on multimodal indicators that capture complex student interactions, enhancing the efficiency and effectiveness of educational tools and environments \cite{wise2021collaborative, laal2012benefits}. Secondly, the use of multimodal indicators via LCA can improve stakeholder communication and decision-making. School administrators, policymakers, and educational technology developers can gain clearer, more actionable insights into student learning processes, aiding in the design of more effective collaborative activities \cite{zhao2024epistemic, kasepalu2021teachers}. This supports a more evidence-based approach to educational policy and curriculum design, leading to improved student outcomes and more inclusive learning environments \cite{florian2011exploring}. Finally, this methodology aligns with learner-centred education principles by recognising and adapting to the diverse behaviours and needs of students. By identifying distinct latent classes of behaviour, educators can provide targeted support, fostering a more supportive and engaging learning atmosphere essential for student motivation and success \cite{urdan2006classroom}.

While this study demonstrates the benefits of integrating LCA within MMLA, several limitations need attention. The current findings illustrate the potential of using LCA to categorise monomodal into multimodal indicators. However, a broader study with diverse datasets is needed to understand the approach’s wider applicability. Determining the optimal number of latent classes is challenging and often subjective. Relying solely on statistical criteria like BIC and log-likelihood values can misalign with theoretical considerations, leading to over- or underestimation of classes. Future studies should align latent classes with specific learning scenarios and theoretical foundations. LCA assumes each behaviour fits neatly into predefined classes, potentially oversimplifying interactions. Exploring methods like hidden Markov models or deep learning could offer more nuanced insights in dynamic learning environments. Finally, while LCA streamlines monomodal indicators into multimodal ones, these resultant indicators can still be challenging to interpret, especially when spanning multiple modalities. Future research should determine if such refinements aid or hinder practical stakeholder interpretation.
% \section{Conclusion}
\begin{acks}
This research was in part supported by the Australian Research Council (DP220101209, DP240100069) and Jacobs Foundation. L.Y.'s work is fully funded by the Digital Health CRC (Cooperative Research Centre). D.G.'s work was, in part, supported by the DHCRC and Defense Advanced Research Projects Agency (DARPA) through the Knowledge Management at Speed and Scale (KMASS) program (HR0011-22-2-0047). The DHCRC is established and supported under the Australian Government's Cooperative Research Centres Program. The U.S. Government is authorised to reproduce and distribute reprints for Governmental purposes notwithstanding any copyright notation thereon. The views and conclusions contained herein are those of the authors and should not be interpreted as necessarily representing the official policies or endorsements, either expressed or implied, of DARPA or the U.S. Government.
\end{acks}

\bibliographystyle{ACM-Reference-Format}
\bibliography{0_reference.bib}

% \appendix





\end{document}